# Morphology-aware sample assignment: overcoming IoU insensitivity for surface defect detection

Pengfei Liu[1,2,†], Yuanbo Wang[4,†], Deng Zhang[4,†], Changguang Song[5,†], Biwei Wu[1,2,*], Fang Li[6,*], Weibo Wang[1,2,3*]

[1] *Center of Ultra-precision Optoelectronic Instrument Engineering, Harbin Institute of Technology, Harbin 150001, China*
[2] *Key Lab of Ultra-precision Intelligent Instrumentation, (School of Instrumentation Science and Engineering, Harbin Institute of Technology), Ministry of Industry and Information Technology, Harbin 150001, China*
[3] *Zhengzhou Advanced Research Institute of Harbin Institute of Technology, Zhengzhou 450000, China*
[4] *Xi'an Institute of Applied Optics, No. 9, West Section of Electronic Third Road, Xi'an 710065,China*
[5] *Research Institute of Network Communications, Shijiazhuang 050051, China*
[6] *College of Computing and Data Science, Nanyang Technological University, Singapore 639798*
[†] *Co-first authors*
[*] *Corresponding Authors*

Intersection-over-Union (IoU), as a pivotal metric for evaluating the spatial alignment between candidate proposals and ground-truth annotations, directly determines the quality of positive sample sets and the training efficacy of visual detection models. Through theoretical modeling and analysis, we uncover a non-sensitive region on the IoU response curve, within which samples yield nearly identical IoU scores despite distinct geometric overlaps. To overcome this limitation, we introduce a set of morphological similarity metrics covering area, shape, and aspect ratio, to refine the positive sample assignment process, thereby ensuring more discriminative and reliable matching. A supplementary matching score is derived via mean-based aggregation of these multidimensional similarities, compensating for the intrinsic limitation of IoU in representing structural correspondence. Theoretically, incorporating morphological similarity reshapes the response distribution of the matching function, yielding both effective directional gradients and polygon-like iso-response contours, which tightly confine high-response regions around each ground-truth instance and substantially enhance the precision of positive sample selection. Experiments based on the YOLOv9 framework demonstrate consistent performance gains on both NEUDET and GC10-DET datasets. Notably, the proposed approach is fully plug-and-play and incurs zero additional inference overhead, thereby ensuring deployment efficiency for industrial visual inspection.

*Codes are released at (https://github.com/PanffeeReal/MCC).*

## 1. Introduction

With the rapid evolution of deep learning theory, visual inspection has become a key enabling technology in modern industry, quality control, and production line monitoring [1,2]. Unlike generic object detection tasks that emphasize large-scale datasets and high-level semantics, industrial surface defect detection demands fine-grained geometric alignment and high-precision localization, where multi-scale and morphologically diverse defects pose significant challenges to detection accuracy and sample assignment reliability.

In contemporary detection frameworks including transformer-based architectures, the IoU metric remains the core criterion for evaluating geometric correspondence between candidate boxes and ground-truth annotations. However, our theoretical modeling reveals an intrinsic limitation of this paradigm: the IoU response curve exhibits a non-sensitive region, within which multiple candidate boxes produce nearly identical IoU values, making it difficult to distinguish subtle geometric differences. This insensitivity compromises the accuracy of positive sample selection and undermines the overall training efficacy of industrial detection models.

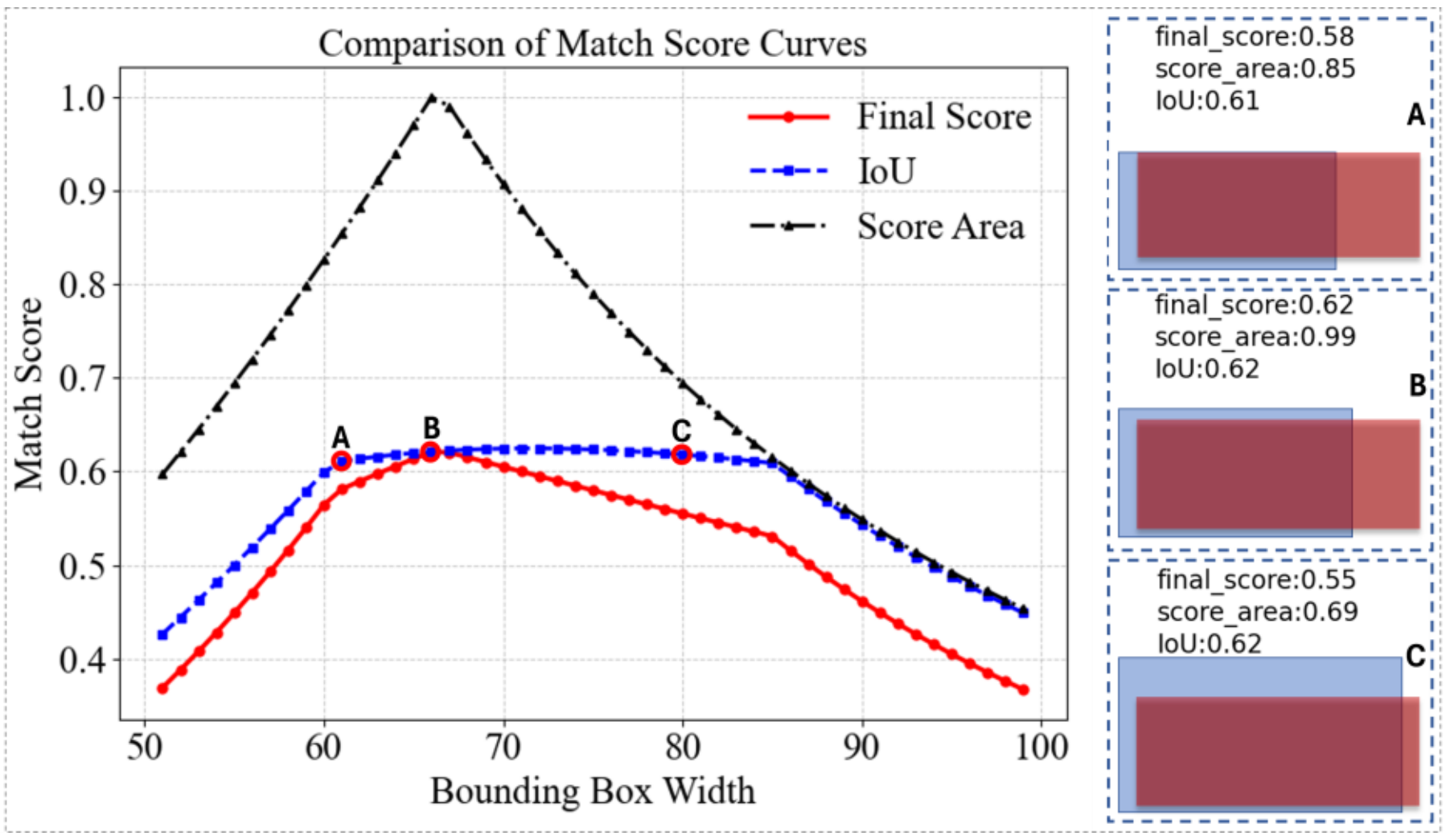

Fig.1 Comparison of matching score curves: IoU, area and the final matching scores are colored as blue, black and red. A, B and C are samples from IoU's insensitive regions. Overlaps and match scores are listed in the right part, where ground-truth and candidate box are colored as red and blue.

As presented in **section 3**, the candidate boxes surrounding a ground-truth instance can be parameterized by (*x,y,wh_ratio,area*). Based on this representation, we constructed a candidate box space and analyzed its matching score distributions. Fig.1 compares the matching score curves of a sampled branch of candidate boxes, which clearly reveals a non-sensitive region on the IoU response curve, a plateau region where multiple geometrically distinct boxes, such as A, B, and C, yield nearly identical IoU values (≈0.62) despite clear differences in overlap. This plateau behavior leads to ambiguous sample ranking and weak gradient propagation during optimization. Consequently, relying solely on IoU hinders the model's ability to accurately distinguish closely distributed samples with subtle geometric variations, thereby impairing the precision of positive sample selection and the efficiency of model training.

We further provide a theoretical visualization in the Appendix illustrating an ideal IoU response curve without a non-sensitive region, which rarely occurs when the candidate box and ground-truth share coincident adjacent edges, an extremely low-probability condition. However, in the vast majority of realistic cases, the endpoints of candidate boxes do not coincide with those of the ground truth. Consequently, the existence of a non-sensitive region in the IoU curve inevitably causes assignment ambiguity: samples with nearly identical IoU scores but distinct geometric alignments are indistinguishably grouped into the same positive sample set. This issue becomes more pronounced when multiple ground-truth instances are spatially close or overlapping, where IoU-based similarity metrics may lead to cross-instance assignment errors.

To overcome this limitation, we establish a morphology-aware sample quality evaluation framework that jointly leverages low-level geometric attributes, characterized by area, aspect ratio, and shape descriptors. This design explicitly decouples geometric similarity into multiple interpretable components, enabling precise differentiation among dense candidates. Consequently, the proposed formulation enables fine-grained distinction among closely distributed candidates that would otherwise yield similar IoU scores. This enhanced representation effectively improves the precision of positive and negative sample separation, leading to more data-efficient model training. In **Section 4**, we conduct a series of simulation analyses on the variation curves of each morphological feature and theoretically demonstrate their capability to alleviate the insensitivity region in IoU matching. As exemplified in Fig.1, the area-based similarity curve successfully mitigates the flat segment present in the original IoU curve.

Building upon this foundation, the proposed Morphological Characteristic Cost (MCC) feature selection mechanism enhances the discriminative capability of sample quality evaluation across multiple dimensions. It effectively resolves IoU's intrinsic insensitivity region, which forms blind zone in reliable candidate evaluation. By incorporating multi-dimensional similarity measures, MCC reshapes the matching score landscape such that the candidates' scores around each ground-truth exhibit smooth and directional gradients. This refinement compresses high-response regions and eliminates flat plateaus, thereby enabling more accurate fine-grained discrimination among densely distributed samples. Consequently, positive samples are more tightly constrained within the immediate vicinity of each ground-truth, forming a compact and reliable positive set. The proposed approach proves particularly effective for industrial surface defect detection, where defect scales and morphologies vary drastically. Extensive experiments on the YOLOv9 family demonstrate consistent accuracy improvements with zero additional computational or inference overhead.

The main contributions and novelties of this work are summarized as follows:

**a)** We construct a parameterized modeling framework of candidate boxes, enabling theoretical and visual analysis of the IoU's insensitivity region. The analysis reveals that IoU's insensitivity region inherently limits the detector's ability to accurately match multi-scale candidates and ground-truth instances, leading to potential cross-instance assignment error and degraded training efficiency. These findings provide the theoretical motivation for our proposed solution.

**b)** We propose a set of morphological similarity measures called MCC, which jointly evaluate area, shape, and aspect-ratio descriptors. These complementary metrics facilitate multi-scale and multi-morphological evaluations of candidate–instance similarity, yielding smooth and directional gradients in the matching response heatmap over the width–height plane. This refinement effectively eliminates the blind zones inherent in IoU metric and substantially enhances both the sensitivity and accuracy of similarity computation.

**c)** The proposed method exhibits strong interpretability. Multi-dimensional similarity formulation reshapes the matching response landscape, whose iso-response contours acquire polygonal structures while the high-value region becomes more compact and localized. Consequently, positive sample selection is tightly constrained to the neighborhood of each ground-truth box, thereby reducing false positives. Along with the multi-dimensional fine-grained distinction among closely distributed samples, the precision of positive sample set can be continuously enhanced, which mitigates assignment error and enables data-efficient training.

**d)** Benefiting from its plug-and-play design, the proposed MCC mechanism can be seamlessly integrated into YOLOv9 series models. Experimental results on NEUDET and GC10-DET datasets validate that optimizing positive sample selection leads to consistent and reliable performance improvements without compromising inference speed, demonstrating strong potential for deployment in real-world industrial inspection systems.

## 2. Literature review

Accurate identification of task-relevant samples plays a crucial role in improving both the learning efficiency and detection accuracy of visual models. Early detectors [3] adopted IoU-based feature selection strategies, assuming a positive correlation between the IoU score and sample quality. In these frameworks, candidate boxes with IoU values exceeding a preset threshold were designated as positive samples. IoU was also extended to supervise localization task, giving rise to a series of optimization functions such as CIoU, GIoU [4,5]. To alleviate the complexity introduced by predefined anchors, anchor free detectors such as FCOS [6] were subsequently proposed. They treat feature points as candidates and treat spatial relationship between the feature location and the ground-truth center as the similarity metric, thereby eliminating anchor-related hyperparameters and simplifying the detection pipeline. Subsequent advanced models [7] combined multiple geometric cues such as angles to improve positives selection.

Despite these advances, the fix rule assignment strategies remain inherently coarse-grained, since the positive sample sets are static and independent of the model's evolving learning dynamics. To address this limitation, dynamic label assignment methods have been proposed. ATSS[8] proposed a dynamic IoU threshold by statistically analyzing the mean and variance of IoU scores across feature levels. TOOD [9] leveraged classification scores to optimize the sample similarity computing and selects samples with the top-k highest scores, thereby improving the quality of positives selection. Given that hard assignment strategies are insufficient to handle the complex correspondence between feature and instances, [10] formulated label assignment as an optimal transport problem, defining a unit transportation cost to assign samples with weighted correspondences to multiple ground truths. TGFS [11] has made progress in feature selection strategies by characterizing feature quality based on mean loss and proposing a dynamic, adaptive feature selection and weight optimization strategy. More recently, transformer-based detectors have achieved end-to-end detection through one-to-one label assignment, eliminating the need for NMS post-processing [12]. However, their similarity evaluation of label assignment is still IoU-based or improved matching costs combining classification and localization terms [13,14].

Through theoretical modeling and analysis, this study reveals the inherent limitation of IoU-based sample assignment in accurately distinguishing ambiguous samples. Due to the intrinsic insensitivity region, within which IoU fails to provide effective gradients, its ability to separate hard examples is weakened. This issue becomes particularly critical in industrial defect detection, where surface defects exhibit high diversity and unpredictability. To address this, we propose a multidimensional similarity modeling approach based on visual morphological features, which theoretically explains how low-level visual cues contribute to improving the quality of sample–instance matching. By reshaping the response distribution of the similarity function, our method ensures effective gradients in arbitrary directions and compresses the high-response region, thus improving the precision of positive sample selection and enhancing training efficiency without introducing any additional computational cost.

Moreover, our theoretical analysis also provides an explanatory perspective for existing multi-task collaborative optimization methods, which incorporate classification scores as additional matching cues to alleviate the insensitivity problem of IoU.

However, unlike our morphologically grounded similarity modeling, classification scores are inherently uncertain and easily influenced by false-positive predictions, which may compromise the accuracy of sample assignment.

## 3 Candidate Box Parameterization and IoU Insensitivity

IoU, as a crucial parameter in detection frameworks, is not only used to evaluate the localization accuracy of predicted boxes but also serves as a key criterion for positive and negative sample assignment during training. In contrast to conventional coordinate-based formulations, we parameterize the candidate box space to enable theoretical and visual examination of IoU's distribution characteristics, particularly its insensitivity region. Given a ground-truth box, its surrounding candidate boxes $B_{\text{can}}$ can be uniquely parameterized as:

$$B_{\text{can}} = (x, y, wh_ratio, area) \tag{1}$$

where *(x,y)* denotes the starting coordinates of $B_{\text{can}}$ (by default, the bottom-left corner), while *wh_ratio* and *area* represents the width-to-height ratio and area. All parameters are non-negative.

We define all candidate boxes sharing the same starting point (*x*,*y*) as a candidate box cluster. Within a cluster, the subset of boxes with the same *wh_ratio* or *area* is referred to as a branch. As illustrated in Fig.2 (a), the visualization of a candidate box cluster anchored at (20,20) shows that, for each equal-*wh_ratio* branch, the endpoints (top-right corners, indicated by circles) of candidate boxes align along a ray (indicated by arrows). Rays corresponding to different *wh_ratio* branches radiate outward from the same origin, forming a fan-shaped pattern. In contrast, Fig.2 (b) visualizes an equal-*area* branch, where the endpoints of candidate boxes trace an arc (the arrow indicates the direction of endpoint variation). Arcs corresponding to different *area* branches extend across the plane in smooth, wave-like trajectories.

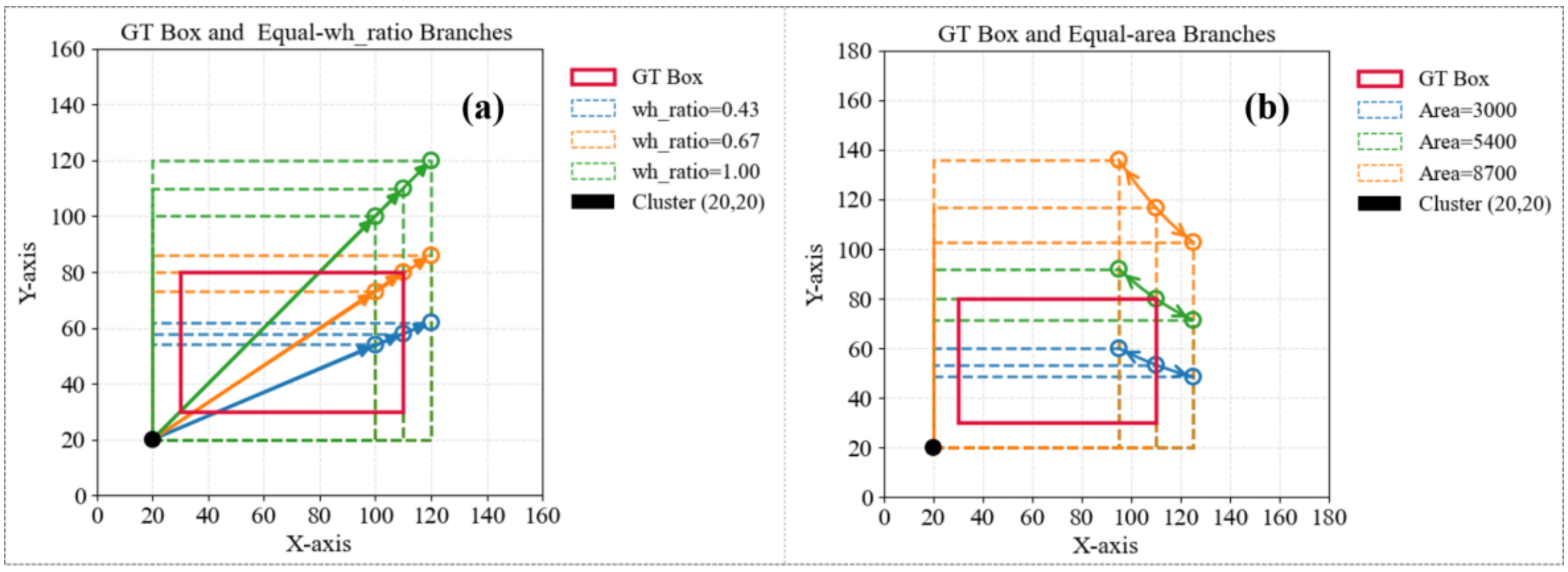

Fig.2 Visualization of different candidate box clusters.

The proposed modeling exhibits **two key properties**. First, it guarantees a one-to-one correspondence between parameters and candidate boxes, ensuring that each unique parameter set defines a single candidate box $B_{can}$. Second, the visualization results demonstrate the completeness of this modeling, as $B_{can}$ can represent all possible candidate boxes within the coordinate space. These properties collectively ensure the comprehensiveness and reliability of the subsequent IoU distribution analyses.

In practical scenarios, the bottom-left corner of the input image is set as the coordinate origin, and the image occupies the first quadrant of the Cartesian plane. All coordinates are computed using pixels as the minimum unit of measurement.

Fig.3 and Fig.4 respectively visualize the IoU score distributions of two distinct branches: equal–*wh_ratio* and equal–*area*, within a candidate box cluster anchored at the same starting point.

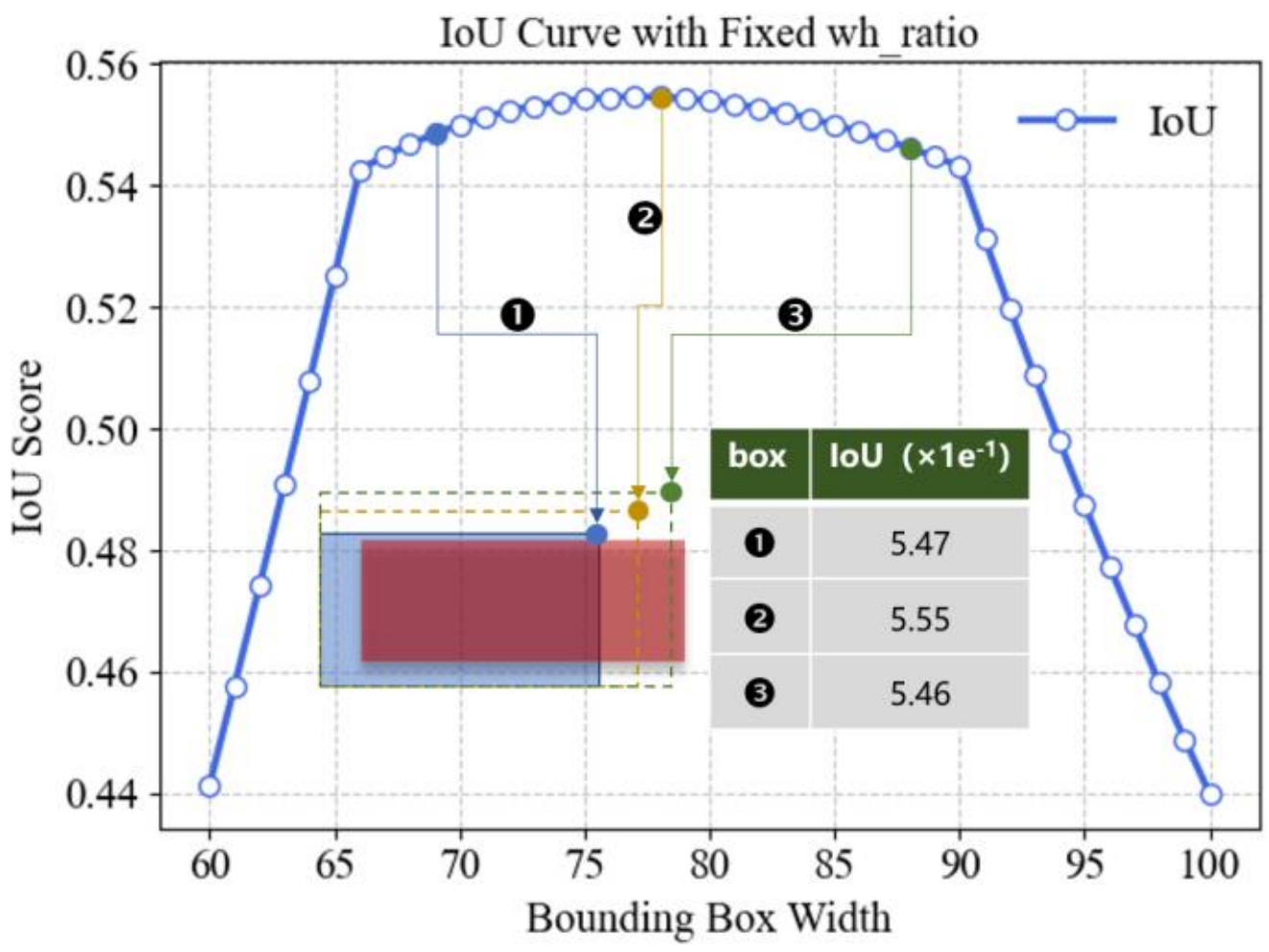

Fig.3 IoU score curves for candidate bounding box sets with the same aspect ratio within the same starting-point candidate cluster. The red background represents the ground-truth box $B_{gt}$=(30,30,1.6,$80^2$/1.6), while the blue curve denotes the IoU scores for the branch box set {$B_{can}$=($x$,$y$,$wh_ratio$,$area$)| $x$=20, $y$=20, $wh_ratio$=1.1, $area$=$(60:1:100)^2$/ $wh_ratio$}. Circles on the curve mark individual sampling points. Furthermore, the blue, brown, and green candidate boxes correspond to three sample points within the non-sensitive region, with their respective IoU scores listed in the accompanying IoU table.

Although the relative starting position of a predicted box with respect to the ground-truth box can take nine possible configurations: combinations of (left, middle, right) and (bottom, middle, top) along the two axes. Our simulations indicate that the IoU curve exhibits a similar variation pattern across all cases. Therefore, only the representative combination (left, bottom) is presented in Fig. 3 and Fig. 4 for illustrative purposes.

As visualized, the IoU curves of two branches can generally be divided into three phases: rapid increase, slow variation, and rapid decrease. For the equal–*wh_ratio* case shown in Fig.3, the rapid increase phase begins when the candidate box first intersects with the ground-truth box. The absolute increment of the intersection area, denoted as $\Delta S_{InSecT}$, can be expressed as the sum of two directional increments and a cross term,

$$\Delta S_{\mathrm{InSecT}} = (a+\Delta x)*(b+\Delta y) - a*b = \Delta x*(b+\Delta y/2) + \Delta y(a+\Delta x/2) \tag{1}$$

where $a$ and $b$ represent the length and width of the intersection rectangle at the previous moment, and Δx and Δy denote the incremental changes along the horizontal and vertical axes, respectively. Although the candidate box area also varies during this process, the ground-truth area remains constant. Consequently, the growth rate of the intersection area exceeds that of the union area, leading to a steep rise in the IoU curve.

The rapid decrease phase begins once the candidate box fully overlaps the ground-truth box. Beyond this point, the intersection area remains constant while the candidate box area continues to increase, resulting in a sharp decline in IoU.

The slow-variation phase occurs when the candidate box and ground-truth box become fully overlapped along one spatial dimension (horizontal in Fig.3). As the candidate box continues to enlarge, increments in intersection area arise only from the other dimension, substantially reducing the rate of change of the intersection area. As a result, the rates of change in the intersection and union areas become approximately proportional, leading to indistinguishable IoU responses for geometrically different samples, the so-called IoU insensitivity region. In the example of Fig.3, this insensitivity region spans a width interval of 66~90 pixels, within which the candidate boxes yield nearly identical IoU scores (approximately 0.55). In this region, IoU alone fails to accurately differentiate among these candidate boxes.

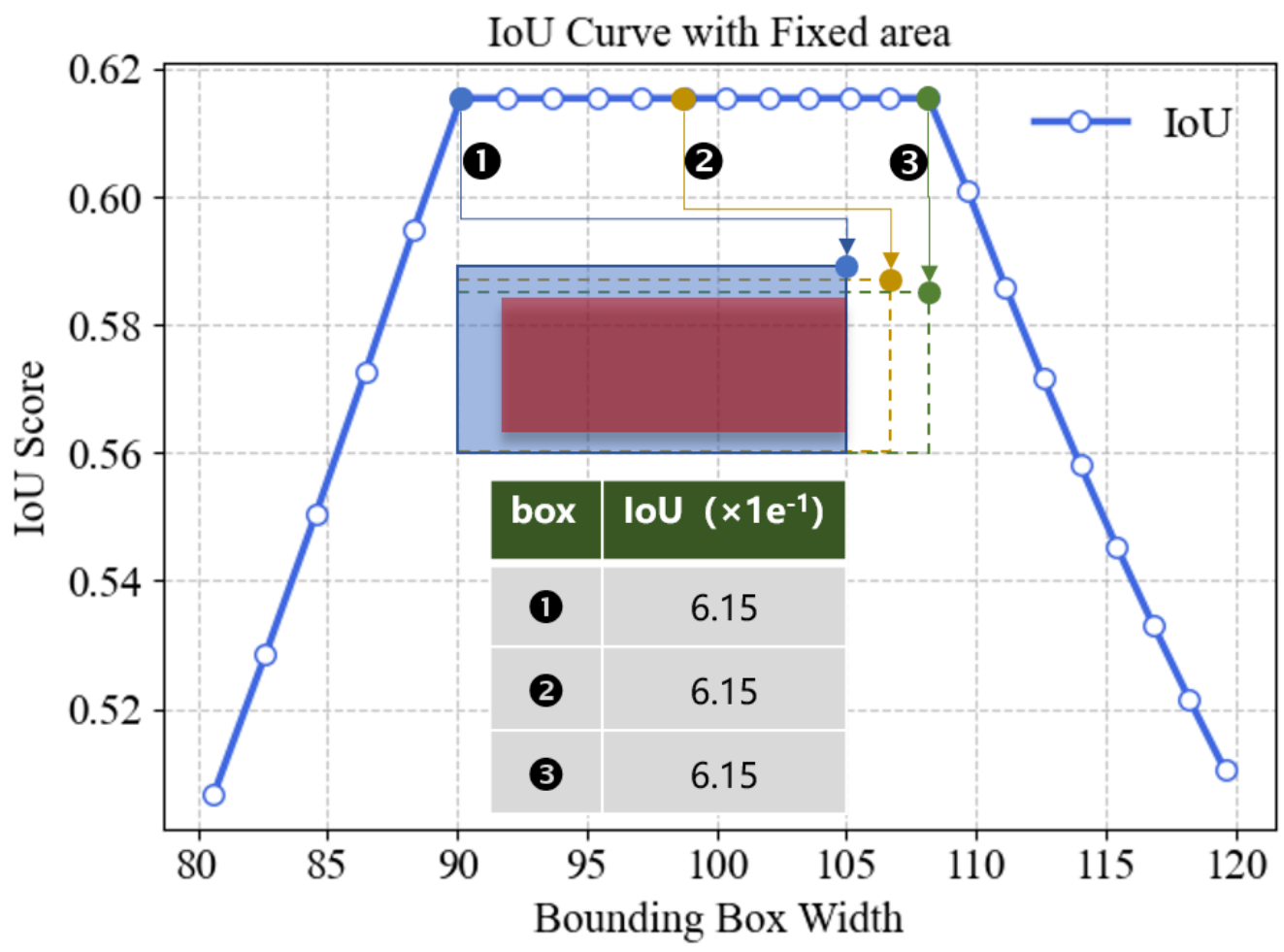

Fig.4 IoU score curves for candidate bounding box set with the same area within the same starting-point candidate cluster. The red background represents the ground-truth box $B_{gt}$=(30,30,1.6,4000), while the blue curve denotes the IoU distribution for the set {$B_{can}$=($x$,$y$,$wh_ratio$,$area$) | $x$=20, $y$=20, $wh_ratio$=1:0.05:2.25, $area$=6500}. The horizontal axis still adopts $width = \sqrt{wh_ratio * area}$ for a more intuitive representation.

For the equal–*area* branch illustrated in Fig.4, both the candidate box and the ground-truth box maintain constant areas throughout the process. Therefore, the IoU variation is solely determined by the change in the intersection area. As the aspect ratio varies, the IoU score exhibits an overall trend of first increasing and then decreasing. In this case, the IoU insensitivity region begins when the candidate box fully overlaps the ground-truth box and ends when they cease to be fully overlapped along one spatial dimension. In the example of Fig.4, this region spans a width interval of 90~107 pixels, within which all candidate boxes completely cover the ground-truth box and thus share identical IoU values, rendering the IoU-based sample selection strategy ineffective.

This theoretical analysis clearly indicates that relying solely on IoU is insufficient to accurately evaluate the matching degree among all candidate boxes. The existence of IoU insensitive regions inevitably compromises the precision of positive–negative sample assignment and degrades the training efficiency of detection models.

To overcome this limitation, we further exploit the multi-scale and morphological diversity of surface defects. By modeling multi-dimensional samples representations through similarity computations in scale and shape, the proposed method effectively compensates for the inherent insensitivity of the IoU metric. This design improves the accuracy of positive and negative sample sets and optimizes the model's learning process, while introducing zero inference overhead.

## 4 Morphological Feature-Based Matching Metrics

### 4.1 Area Similarity Metric

In complex and varying detection scenarios, objects inevitably possess multi-scale and multi-morphological characteristics, which is particularly evident in general visual detection and industrial defect inspection. Therefore, we propose to refine the positive sample selection process by calculating the similarity between candidate boxes and ground-truth instances across dimensions such as area and shape, thereby enhancing model training efficiency and detection accuracy.

We first establish the similarity in **scale(size)** based on the area values of the candidate box $Area_{\text{candi}}$ and the ground-truth box $Area_{\text{gt}}$ . Considering the negative correlation between the matching degree and the area difference, we design the scale similarity score as:

$$Score_{Area} = 1 - \frac{\left| Area_{\text{candi}} - Area_{\text{gt}} \right|}{\max(Area_{\text{candi}}, Area_{\text{gt}})} \tag{2}$$

Where $Area_{\text{candi}}$ and $Area_{\text{gt}}$ are the areas of the candidate and ground-truth boxes, calculated as the product of the box's width and height.

The range of $Score_{Area}$ is (0,1], and a larger score indicates higher scale similarity between the candidate box and the ground-truth box. Fig.5 visualizes its response distribution. In the figure, the yellow region, with the red curve $Score_{Area} = 1$ as its spine, represents the set of candidate boxes having a high area matching score with the given ground-truth box (red star). wo white dashed lines illustrate the iso-response contours for two lower matching scores. However, the existence of these contours directly indicates that the area matching score itself suffers from an insensitivity region, making it impossible to further distinguish candidate samples lying on the same contour. This mandates the introduction of new dimensions for similarity evaluation.

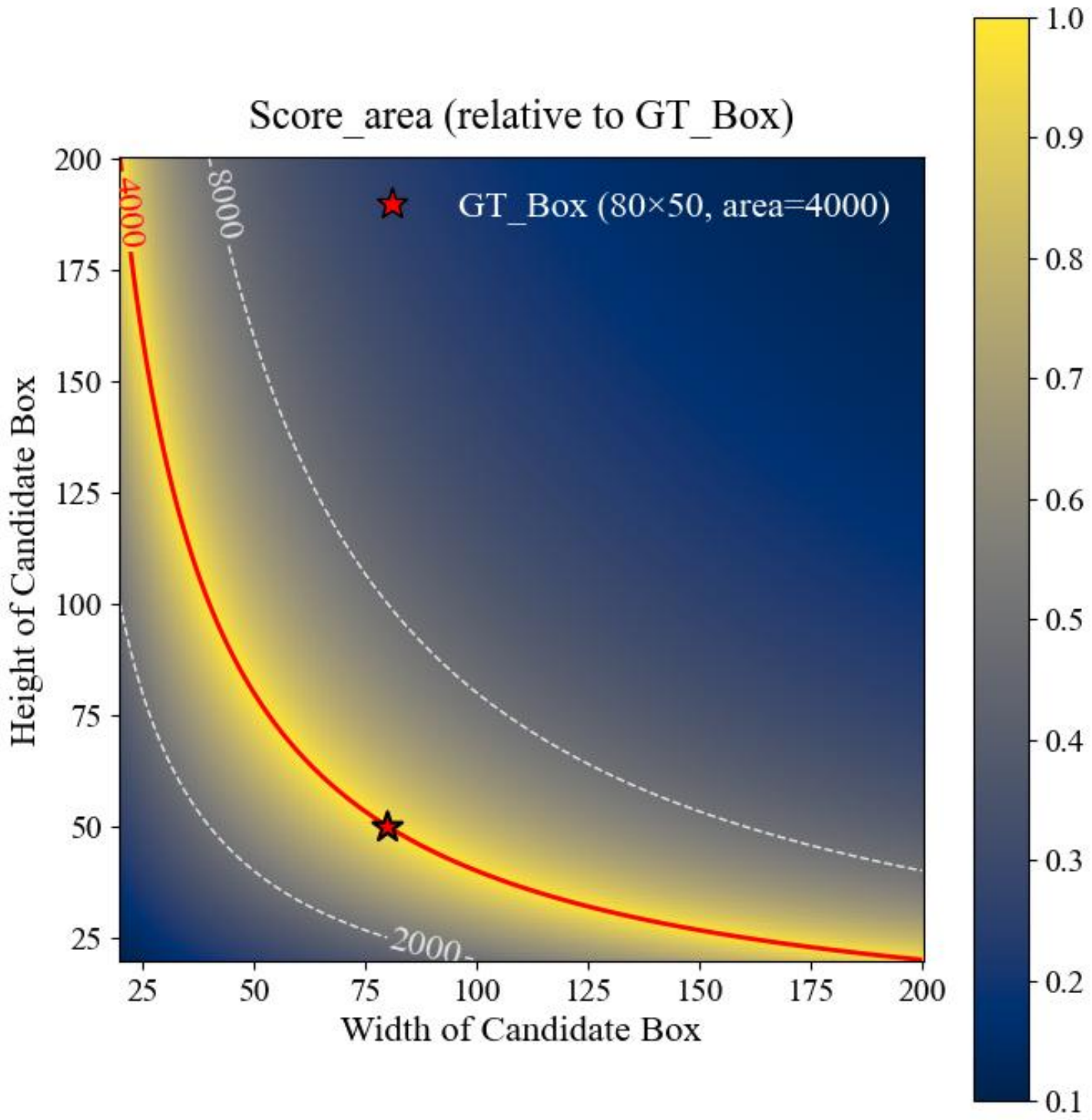

Fig.5 Response distribution of scale similarity.

## 4.2 Shape and Aspect Ratio Similarity Metrics

To compensate for the inherent limitation of the area metric, we subsequently construct a morphological similarity representation based on the **aspect ratio** of the candidate and ground-truth boxes. Likewise, based on the negative correlation between the matching degree and the difference in shape, we define the shape similarity score as:

$$Score_{Shape}=1-\frac{\left|Shape_{\mathrm{candi}}-Shape_{\mathrm{gt}}\right|}{\max(Shape_{\mathrm{candi}},Shape_{\mathrm{gt}})} \tag{3}$$

where $Shape_{\mathrm{candi}}$ and $Shape_{\mathrm{gt}}$ are the shape representations of the candidate and ground-truth boxes, calculated as the ratio of the shorter side to the longer side of the box:

$$Shape=\frac{\min(w,h)}{\max(w,h)} \tag{4}$$

$Score_{Shape}$ effectively evaluates the similarity of objects along the dimension of *slimness* or *squatness*. Its range is (0,1], with a higher score indicating greater similarity. However, it is noted that the response heatmap distribution of $Shape$ is symmetrical about the $w = h$ diagon. This symmetry is inherited by the response distribution of $Score_{Shape}$ , as shown in Fig.6, resulting in two iso-response contours for typical non-square ground-truth boxes. This characteristic is clearly unfavorable for accurate candidate sample selection.

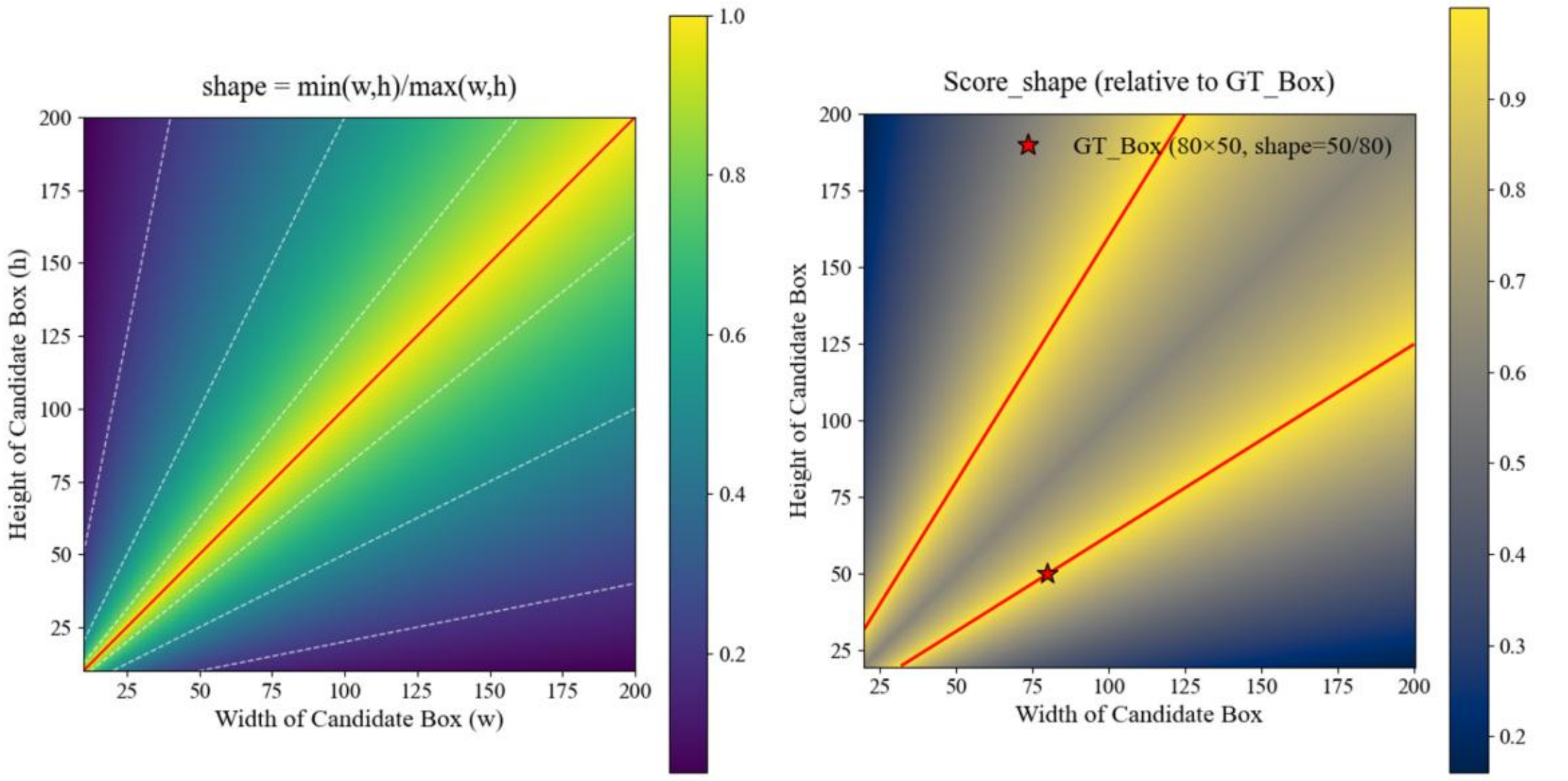

Fig.6 Response distribution of shape similarity.

To eliminate the dual iso-contour phenomenon caused by diagonal symmetry, we finally design the width-height ratio matching score $Score_{WH}$ to further refine the candidate box matching score. This score directly compares the similarity of the candidate and ground-truth boxes in terms of their width and height, as shown in Eqs. (5-6). Its range is (0, 1]. Fig.7 visualizes its response, confirming that it resolves the dual iso-value contour issue caused by symmetry by directly computing similarity in the width-height dimension.

$$Score_{WH} = \frac{\min(Ar_{\mathrm{candi}}, Ar_{\mathrm{gt}})}{\max(Ar_{\mathrm{candi}}, Ar_{\mathrm{gt}})} \quad (5)$$

$$Ar = w / h \quad (6)$$

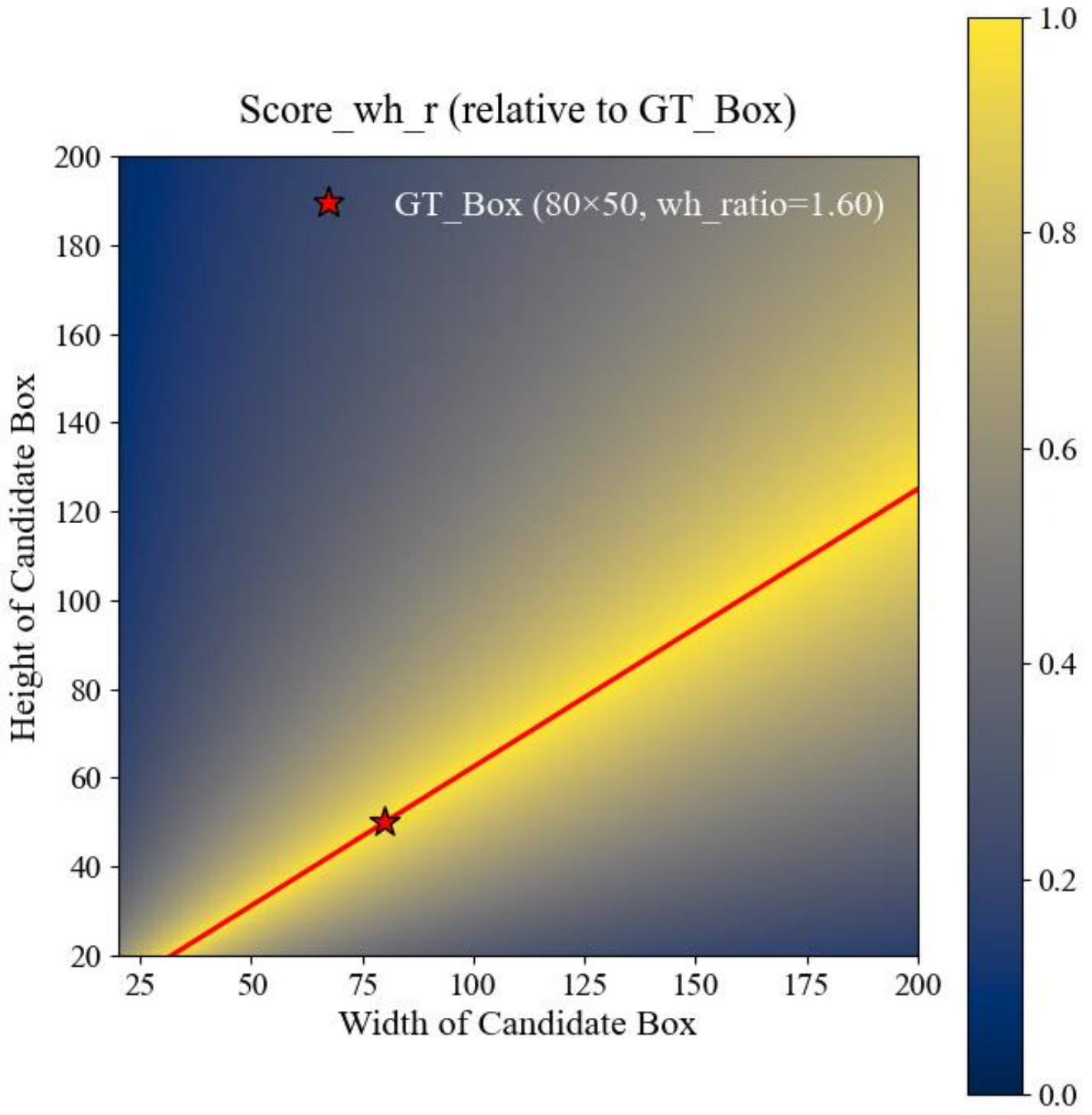

Fig.7 Response distribution of ratio similarity.

### 4.3 Final Morphological Characteristic Cost (MCC)

We adopt a simple averaging summation to generate the final optimized matching metric score, as shown in Eq.(7):

$$Score_{MCC} = \frac{1}{3} \cdot (Score_{Area} + Score_{Shape} + Score_{WH}) \tag{7}$$

Theoretically, by combining the response curves from multiple morphological feature dimensions, $Score_{MCC}$ effectively eliminates the insensitivity region inherent in a single matching metric. Fig.8 visualizes this combined response, showing that the iso-response contours of the single scores are eliminated, replaced by a total response heatmap that exhibits smooth, directional gradients (change in brightness) across all regions.

Upon closer inspection, the overall response still forms three distinct bright ridges corresponding to the three feature directions: Area, Shape, and Width-Height Ratio. Between these ridges, varying gradient regions (slopes) are formed. For instance, the position with parameters identical to the ground-truth (marked with a red star) is the overall similarity response peak. Candidates sharing only the same area or same shape (but different other parameters) form their own local high-response regions, signifying high similarity to the ground-truth in a specific morphological feature dimension.

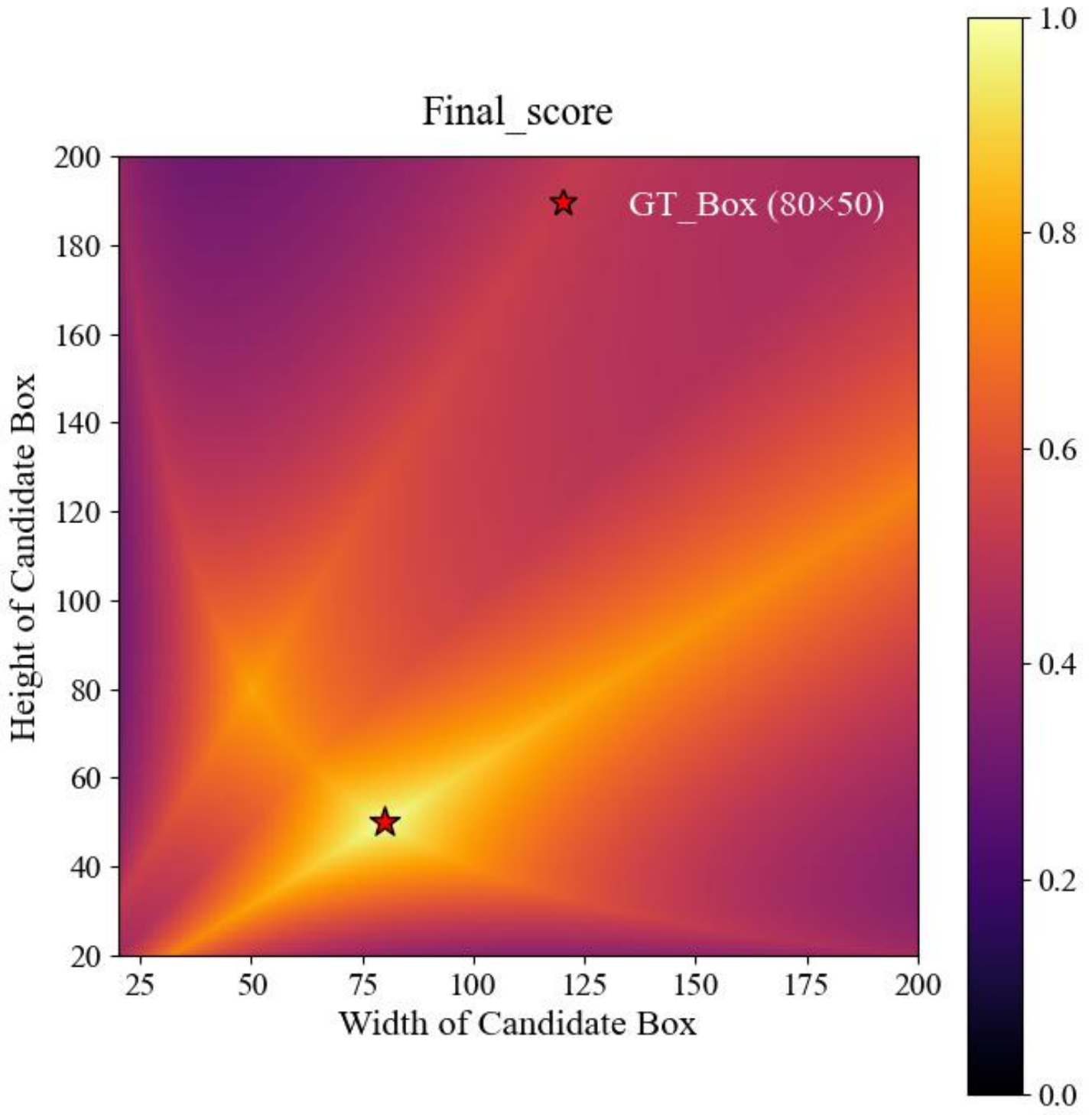

Fig.8 Response distribution of the proposed MCC similarity.

Finally, we define the overall similarity between a candidate sample and the ground-truth instance as the product of the optimized morphological matching metric score and the original IoU score:

$$Similarity = IoU \times Score_{MCC} \tag{8}$$

Eq.(8) comprehensively evaluates the full similarity between the sample and the ground-truth across geometric overlap (IoU) and multiple morphological feature dimensions. This comprehensive metric aids in constructing a more precise positive/negative sample set, thereby enhancing model training efficiency. The core characteristic of our method is the construction of $Similarity$. In practical application, we simply multiply $Score_{MCC}$ with the similarity score used in the training framework's assigner.

Based on the visualized response heatmap of our method, Fig.9 compares the 0.5 iso-response contours used for positive sample selection by the IoU criterion (green) and the proposed $Similarity$ (sky blue). Two significant differences are observed:

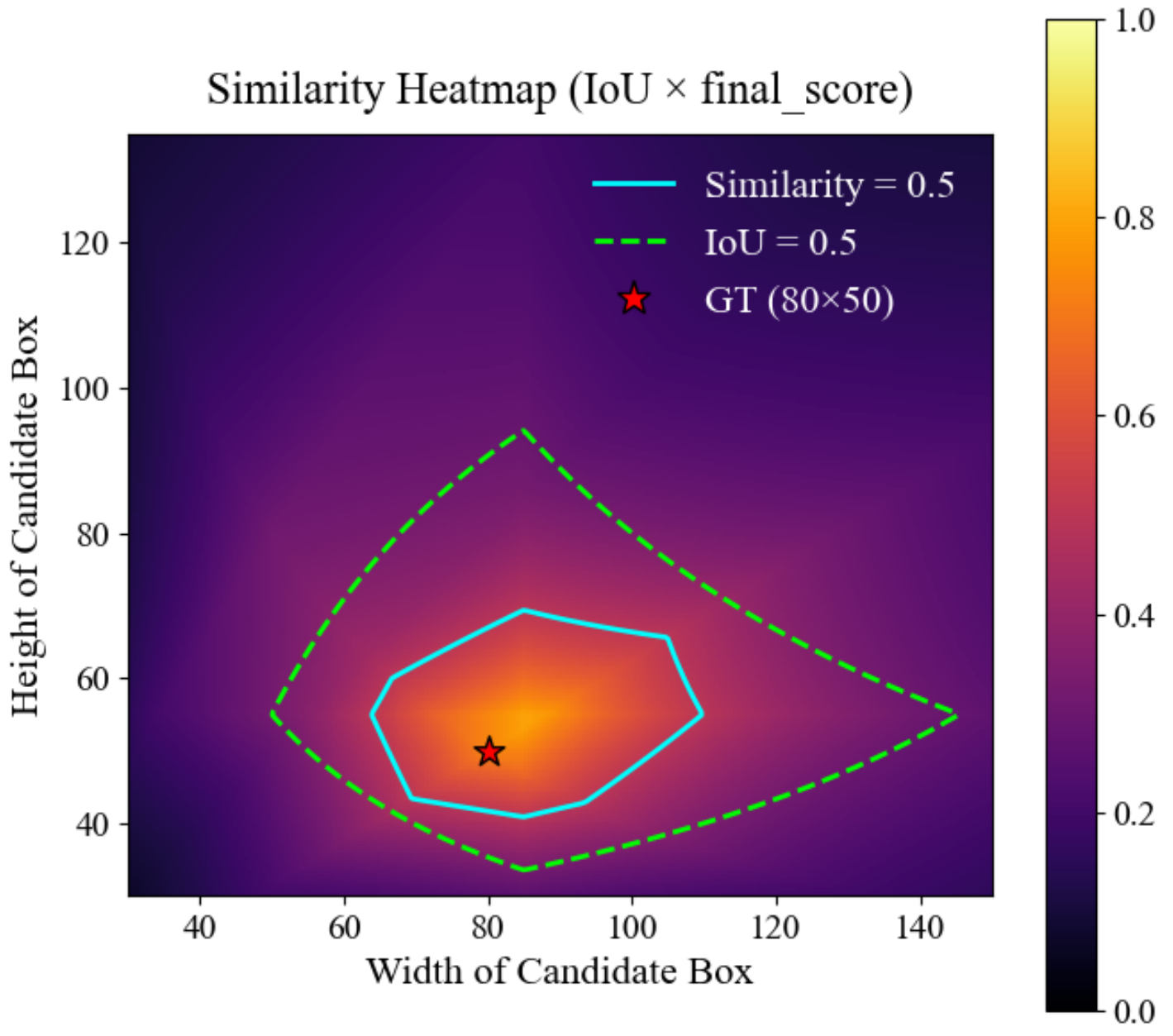

Fig.9. Response heatmap of *Similarity* and comparison of iso-response contours. ($B_{gt}$ = (25,25,1.6,4000)).

**1) Enhanced Positive Sample Compactness:** According to the iso-response contour distribution, the proposed method constrains positive samples primarily to the immediate vicinity of the ground-truth box, forming a smaller, more precise high-response region. This effectively prevents low-matching candidates far from the ground-truth from being incorrectly assigned as positives. In contrast, the IoU contour covers a wider range, resulting in a significantly more diffused positive sample region that includes many candidates distant from the ground-truth.

**2) Polygonal Contour Shape:** The contours of the proposed method exhibit a polygonal structural feature, indicating that the similarity metric relies not only on overlapping area but also integrates multi-dimensional, morphology-related information. Conversely, the IoU contours appear similar to a rectangular boundary, reflecting its singular judgment mechanism based solely on geometric overlap.

## 5. Experiments

### 5.1 Baseline and Implementation Details

**Baseline:** Unlike most existing methods that focus on improving model structures, such as dedicated feature extraction modules, our proposed approach is centered on data-efficient training schemes. The method requires no structural modifications to the detection model, granting it a plug-and-play characteristic, compatibility with all mainstream visual detection models, and high practicality. By continuously evaluating the similarity between candidate samples and ground-truth instances across multiple feature dimensions during the model iteration process, our method consistently optimizes the quality of the positive sample set, thereby effectively enhancing the model's learning efficiency and detection performance.

Considering state-of-the-art performance and high deployment potential, we select the YOLOv9 series framework as our baseline to validate the generality and effectiveness of the proposed method. We do not make any changes to the backbone or head structures of the YOLOv9 series code, we only optimize the similarity calculation function within the assigner to incorporate our proposed approach. The same principle applies to other detection models, requiring only the modification of the sample similarity calculation function within their respective assigner tasks. For broader reproducibility and application, our code is publicly available.

**Dataset**: we use the publicly available GC10-DET and the NEU-DET datasets to evaluate the performance of the proposed methods. The GC10-DET dataset comprises 2294 images, including ten types of defects: 1_chongkong (punching, **Pu**), 2_hanfeng (weld line, **Wl**), 3_yueyawan (crescent gap, **Cg**), 4_shuiban (water spot, **Ws**), 5_youban (oil spot, **Os**), 6_siban (silk spot, **Ss**), 7_yiwu (inclusion, **In**), 8_yahen (rolled pit, **Rp**), 9_zhehen (crease, **Cr**), 10_yaozhe (waist folding, **Wf**). The training and evaluation set ratio for GC10-DET is 9:1. The NEU-DET dataset contains a total of 1800 images, equally divided into six kinds of steel surface defects: crazing (**Cr**), inclusion (**In**), patches (**Pa**), pitted-surface (**Ps**), rolled-in-scale (**RiS**), and scratches (**Sc**). The training and evaluation set ratio of NEU-DET is 8:2.

**Implementation Details**: the experimental environment is built with the open-source platform PyTorch and Ubuntu 18.04 operating system. A training-from-scratch strategy is applied to avoid the discrepancy between source and target domains. The input resolution is $512^2$ and $224^2$ for GC10-DET and NEU-DET datasets, respectively. Other training hyperparameters are presented in Table 1.

Table 1. Parameter setting

| **Parameter** | **Value** |
|---|---|
| optimizer | SGD |
| learning_rate | 0.01 |
| momentum | 0.937 |
| weight_decay | 5e-4 |
| warmup_epochs | 3.0 |
| warmup_momentum | 0.8 |
| warmup_bias_lr | 0.1 |
| batch_size | 16 |

**Evaluation:** We use the widely recognized mean average precision (mAP) for accuracy evaluation, calculated as the average of AP values across all categories with an IoU threshold of 0.5. The parameters (Param.) and floating-point operations (Flops) are used to evaluate the model complexity. The number of parameters and Flops of each model were measured using the ptflops library based on the PyTorch implementation. To ensure consistency with the actual inference process, all models were evaluated in the fused inference mode (convolution and batch normalization layers were merged) and set to evaluation state. The calculations were performed on the DetectMultiBackend.model structure with an input resolution of 512×512 (GC10-DET) and 224×224 (NEU-DET). Each multiply–accumulate operation was counted as two floating-point operations. This setting allows a fair and reproducible comparison of computational complexity across different models.

### 5.2 Vertical comparison

Benefiting from the plug-and-play convenience of the proposed method, we compared the performance results of the entire YOLOv9 model series, both before and after applying our method, on the GC10-DET and NEU-DET metal surface defect detection datasets.

Table 2 summarizes the comparative results on the GC10-DET dataset. The overall data in the mAP column confirms the strong generalizability of the proposed method, providing stable performance gains for every model in the YOLOv9 family, with an average net mAP improvement of 2.38% (the average net gain for the 5 models). For instance, the mAP for the small YOLOv9-t detector improves by 2.7% while maintaining constant parameters. For the largest YOLOv9-e model, the proposed method directly elevates the mAP value to 74.2% (a net absolute gain of 2.5%), demonstrating its strong practical application prospects. Regarding computational efficiency, a model-by-model comparison of parameters (Param.) shows that our method introduces zero additional computational overhead or compromised detection speed, which is a significant advantage in practical engineering applications.

Table 2: Comparison of yolov9 series on GC10-DET

| **Model** | **mAP** | **Param.** | **FLOPs** | **AP** (%) | | | | | | | | | |
|---|---|---|---|---|---|---|---|---|---|---|---|---|---|
| | (%) | | (G) | Pu | Wl | Cg | Ws | Os | Ss | In | Rp | Cr | Wf |
| YOLOv9-t | 70.2 | 2662764 | 7.00 | 94.4 | 92.5 | 97.2 | 80.6 | 68.9 | 66.5 | 34.9 | 28 | 53.7 | 85.6 |
| w. ours | **72.9**$_{+2.7}$ | 2662764 | 7.00 | 93.1 | 93.7 | 91.7 | 82.8 | 73.6 | 61.7 | 36.1 | 47.2 | 60.9 | 88.7 |
| YOLOv9-s | 68.3 | 9750300 | 25.24 | 94.7 | 80.3 | 95.6 | 80.8 | 73.5 | 66.3 | 25.6 | 41.0 | 44.6 | 81.0 |
| w. ours | **70.8**$_{+2.5}$ | 9750300 | 25.24 | 93.9 | 95.7 | 95.0 | 79.8 | 72.6 | 58.6 | 30.3 | 37.4 | 55.0 | 89.6 |
| YOLOv9-m | 69.0 | 32777180 | 84.48 | 95.1 | 94.7 | 94.7 | 82.1 | 78.2 | 62.1 | 27.0 | 35.2 | 53.4 | 67.4 |
| w. ours | **71.6**$_{+2.6}$ | 32777180 | 84.48 | 97.4 | 91.1 | 96.7 | 76.7 | 72.9 | 65.3 | 30.2 | 37.1 | 61.6 | 86.9 |
| YOLOv9-c | 70.8 | 51020348 | 152.56 | 94.5 | 95.3 | 96.7 | 76.1 | 73.2 | 63.8 | 28.7 | 47.5 | 54.4 | 77.6 |
| w. ours | **72.4**$_{+1.6}$ | 51020348 | 152.56 | 94.9 | 89.8 | 93.0 | 79.5 | 69.0 | 60.1 | 30.2 | 68.4 | 67.0 | 72.1 |
| YOLOv9-e | 71.7 | 69421692 | 156.28 | 95.3 | 92.7 | 94.3 | 79.1 | 75.8 | 69.4 | 39.0 | 48.0 | 51.1 | 72.6 |
| w. ours | **74.2**$_{+2.5}$ | 69421692 | 256.28 | 95.4 | 94.6 | 95.7 | 77.7 | 71.7 | 61.2 | 31.4 | 75.9 | 53.9 | 84.6 |

Fig.10 visualizes the data from Table 2 using a bar chart comparison, allowing readers to more intuitively appreciate the positive effect of the proposed method. The blue bars represent the original YOLOv9 series models, and the orange bars represent the corresponding models trained with our method. This visually confirms the compatibility and effectiveness of our approach with detection models of varying sizes.

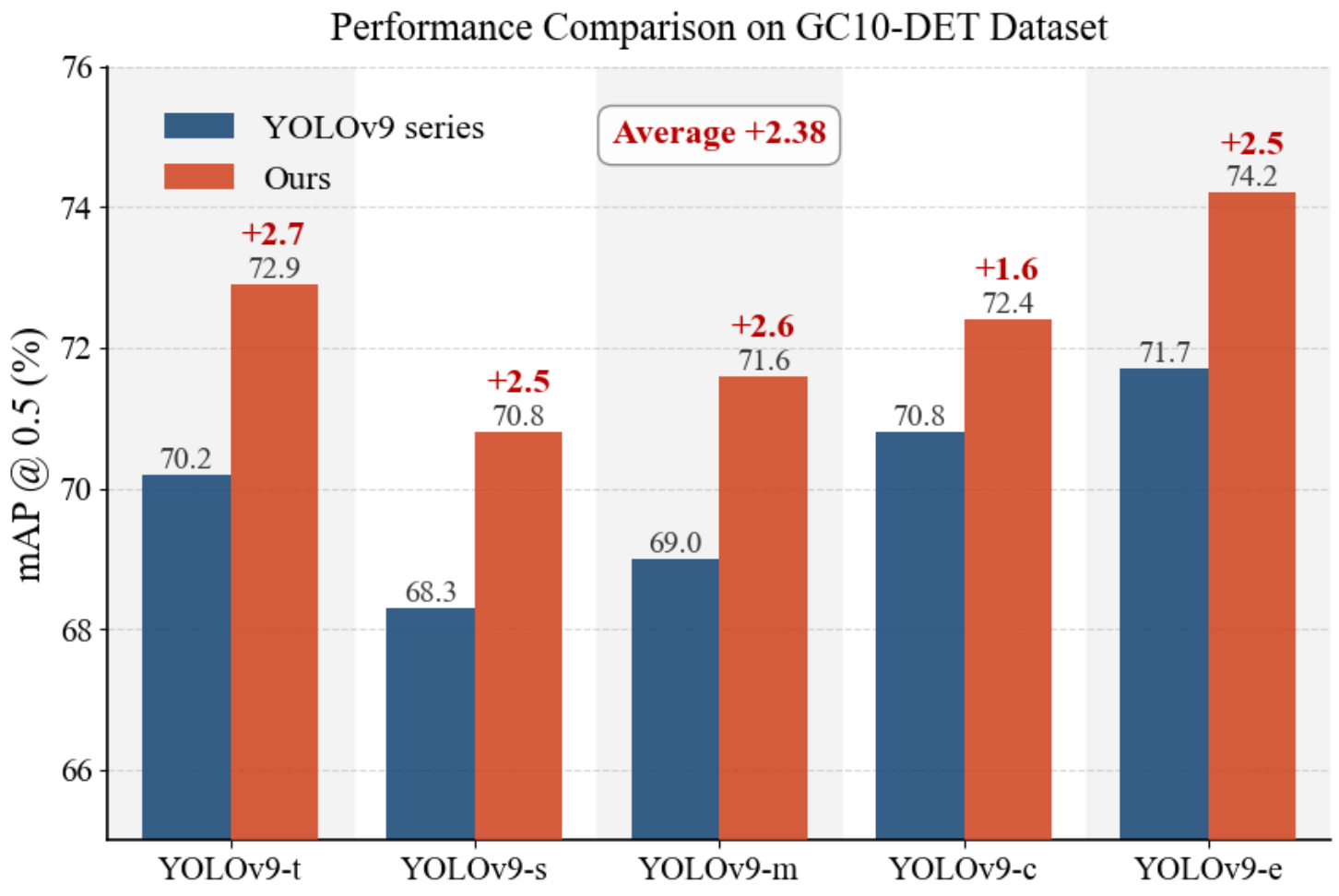

Fig.10 Performance comparison on GC10-DET.

Furthermore, because our method additionally evaluates the similarity between candidate samples and ground-truth instances across multi-scale and multi-morphological dimensions, it effectively enhances the model's learning capability for instances with diverse shapes. A detailed comparison of the AP values for defect types such as Cr (crease) across various models reveals that our approach can significantly boost the detection performance for highly variable morphologies, especially for elongated defects that are typically challenging to detect.

To validate this argument, we employ YOLOv9-m as a representative model due to its optimal balance between detection accuracy and computational complexity. Specifically, we conduct comparative visualizations of feature heatmaps, Precision-Recall (PR) curves, and training dynamics. Fig.11 illustrates the feature heatmaps generated using model parameters before and after the integration of our method, aiming to compare the model's perception of diverse defect types. All detection boxes in the figure are labeled with their respective categories and classification confidence scores.

The results demonstrate that our approach significantly enhances the model's sensitivity and perceptual capability toward multi-scale and irregularly shaped defects. This is evidenced by the high-value regions in the heatmaps, which show superior alignment with the ground-truth defects and exhibit higher activation intensities. Furthermore, this improved perception leads to a substantial gain in the baseline’s recall.

For instance, as shown in the second row, while the baseline model fails to detect two annotated defects, the inclusion of MCC enables successful detection of all instances.

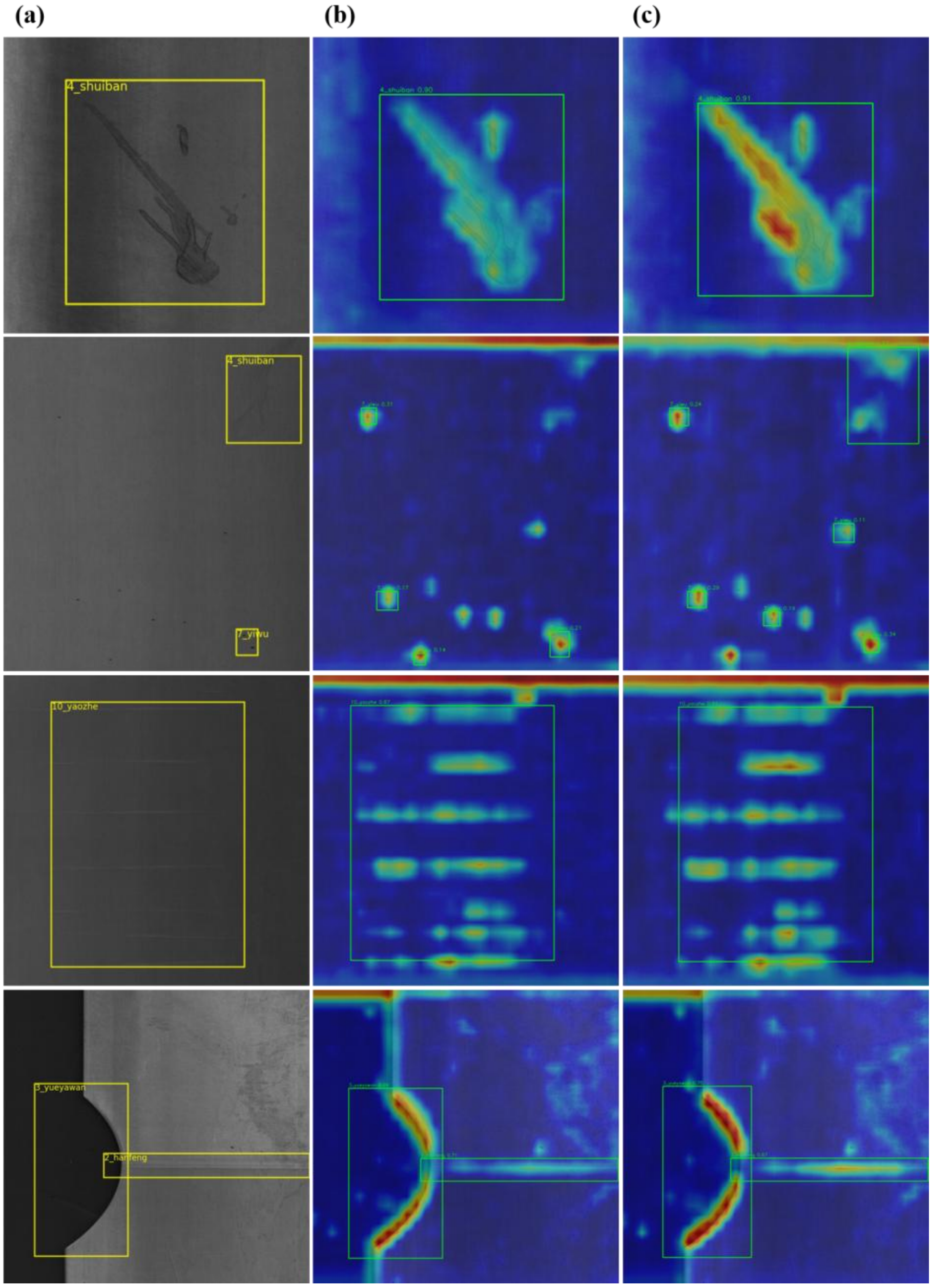

Fig.11 Comparison of detection results and feature heat map: (a) input and annotations, (b) yolov9-m, (c) yolov9-m with MCC.

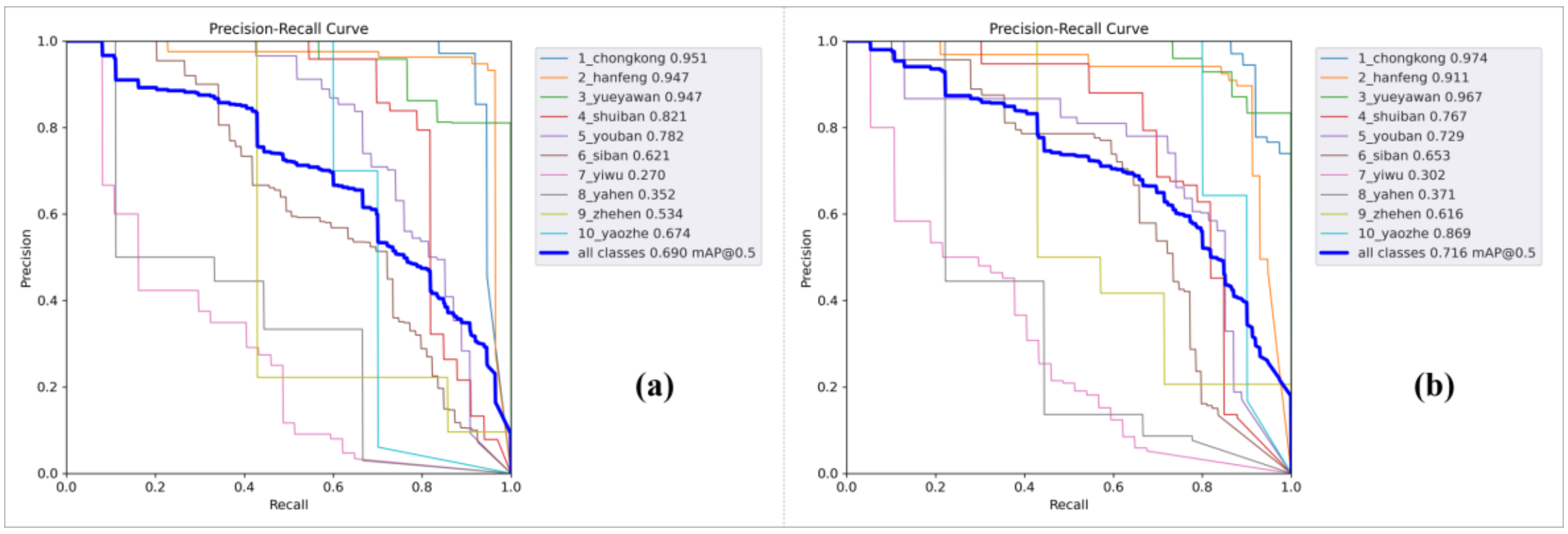

Fig.12 Precision-Recall (PR) curves before and after applying MCC.

Fig.12 illustrates the Precision-Recall (PR) curves before and after the integration of our method. The MCC strategy significantly enhances the detection of challenging defect categories: specifically, the Average Precision (AP) for Category 7 increased by 3.2 percentage points, Category 8 by 1.9, Category 9 by 8.2, Category 10 by a substantial 19.5, Category 6 by 3.2, Category 3 by 2.0, and Category 1 by 2.3. These improvements validate the effectiveness of the proposed method in recognizing defects with diverse morphologies. However, performance decreases were observed in Category 2 (-3.6%), Category 4 (-5.4%), and Category 5 (-5.3%), revealing certain limitations of the current design. Nevertheless, the overall results demonstrate that the proposed approach effectively bolsters the model's capability to identify complex defects in industrial scenarios.

Fig.13 visualizes the training process both with and without our proposed method, highlighting its superior stability and convergence characteristics. As illustrated, the MCC strategy yields more robust trajectories for Precision, Recall, and mAP, characterized by a significant reduction in outliers and fluctuations. Furthermore, our method exhibits accelerated convergence. Specifically, the mAP_0.5 curve enters the high-performance zone at a much earlier stage. While the baseline model requires approximately 80 epochs to surpass the 0.6 mAP threshold, the proposed approach achieves the same performance level by around 50 epoch, representing a substantial improvement in training efficiency.

Collectively, these observations validate that by incorporating the multi-scale and multi-morphological similarity between ground-truth instances and candidate proposals, could effectively enhance the model's feature perception and learning capacity for complex defect targets. Consequently, this leads to superior detection precision without compromising inference speed.

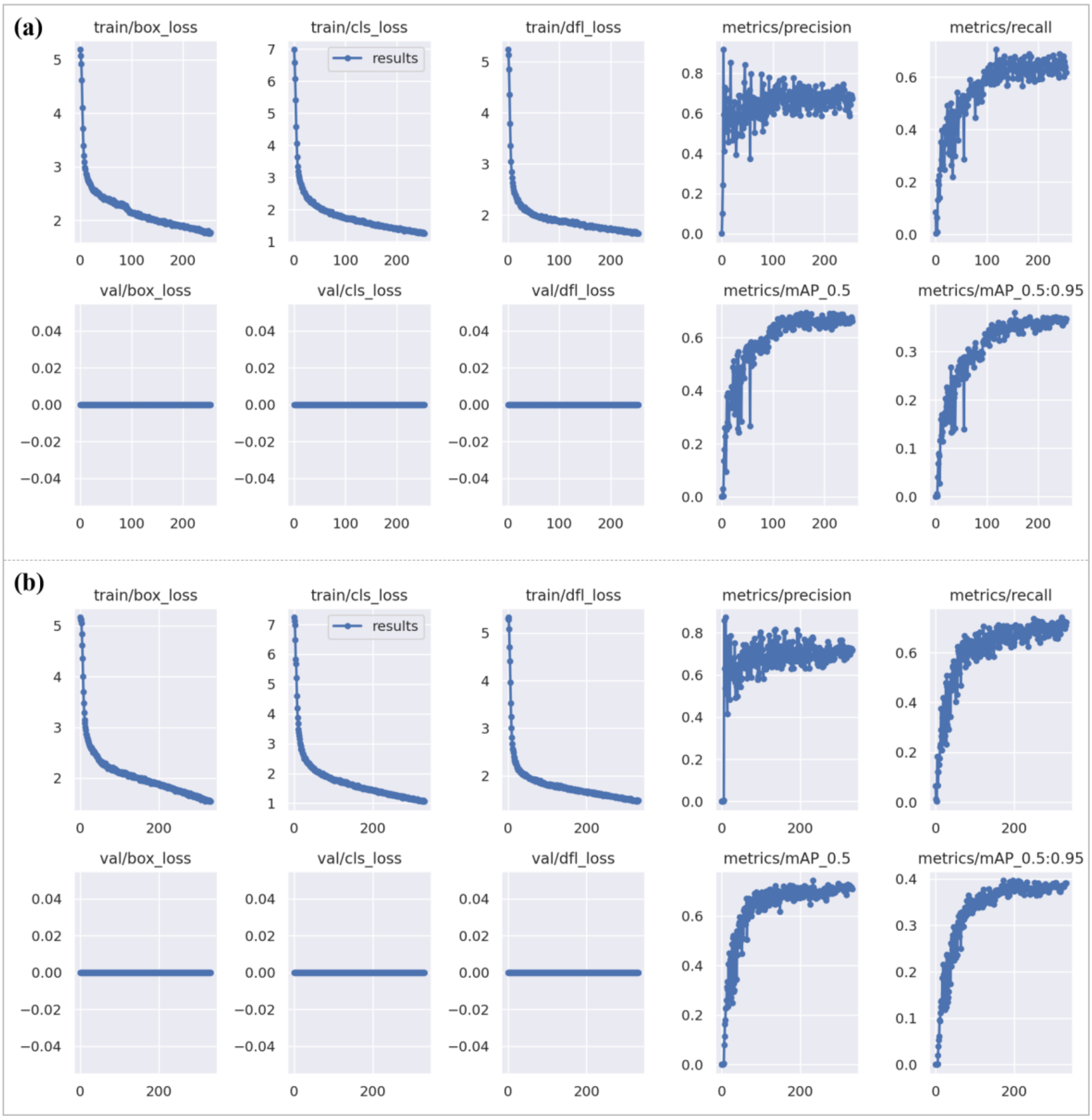

Fig.13 Visualization of YOLOv9-m before (w/o ours, upper (a)) and after (w/ ours, lower (b)) applying the proposed method.

To further verify the generalizability of MCC, we conducted comparative experiments on the NEU-DET dataset. Table 3 summarizes the comparison results. In terms of detection accuracy, the proposed method consistently yields stable improvements in mAP across all YOLOv9 family, notably boosting the mAP of the baseline YOLOv9-t model by 1.2 percentage points. A breakdown by individual model shows that our approach reliably boosts performance at zero cost of increased model complexity.

Table 3. Comparison with yolov9 series on NEU-DET

| **Model** | **mAP** (%) | **Param.** | **FLOPs** | AP(%) | | | | | |
|---|---|---|---|---|---|---|---|---|---|
| | | | | Cr | In | Pa | Ps | RiS | Sc |
| YOLOv9-t | 76.4 | 2661204 | 1.34G | 34.3 | 83.1 | 90.9 | 89.7 | 63.6 | 96.6 |
| w. ours | **77.6**$_{+1.2}$ | 2661204 | 1.34G | 34.8 | 86.9 | 92.8 | 88.3 | 67.6 | 95.1 |
| YOLOv9-s | 76.7 | 9747204 | 4.83G | 33.5 | 87.9 | 90.6 | 87.2 | 64.7 | 96.0 |
| w. ours | **77.7**$_{+1.0}$ | 9747204 | 4.83G | 38.3 | 87.5 | 91.9 | 87.7 | 65.0 | 95.5 |
| YOLOv9-m | 76.9 | 32771396 | 16.17G | 35.1 | 88.7 | 91.7 | 84.8 | 65.9 | 95.4 |
| w. ours | **77.8**$_{+0.9}$ | 32771396 | 16.17G | 39.8 | 85.4 | 91.9 | 87.0 | 66.1 | 96.5 |
| YOLOv9-c | 78.6 | 51011108 | 29.20G | 37.2 | 86.9 | 92.8 | 87.2 | 69.8 | 97.4 |
| w. ours | **79.6**$_{+1.0}$ | 51011108 | 29.20G | 43.3 | 89.7 | 93.1 | 86.8 | 66.9 | 97.8 |
| YOLOv9-e | 78.7 | 69415554 | 29.91G | 40.7 | 84.2 | 92.9 | 87.9 | 69.7 | 96.7 |
| w. ours | **79.8**$_{+1.1}$ | 69415554 | 29.91G | 45.9 | 87.6 | 93.4 | 86.5 | 69.9 | 95.6 |

Fig.14 visualizes the mAP column data from Table 3, providing a more intuitive demonstration of the effectiveness and versatility of the proposed method across different models. Based on the performance gains observed across all models, we calculated that the average mAP improvement for the 6 detectors is 1.08%. Crucially, our method introduces no extra computational parameters or inference speed degradation, distinguishing it from many existing methods and highlighting its potential for practical deployment.

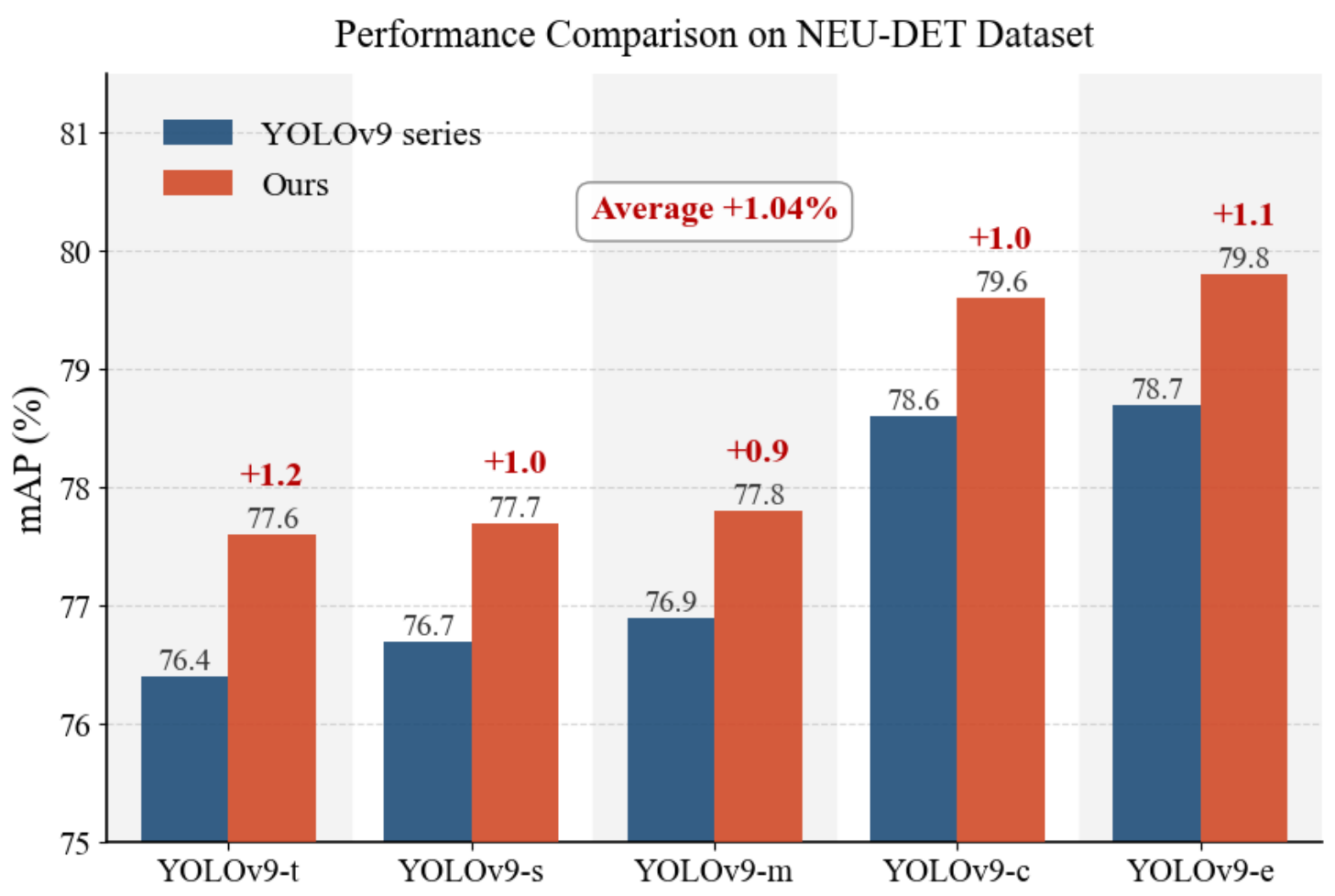

Fig.14. Performance comparison on NEU-DET.

## 5.4 Horizontal comparison

To evaluate the effectiveness and competitiveness of the proposed method, horizontal comparisons were conducted against state-of-the-art general object detectors and specialized defect detection frameworks on two benchmark datasets: GC10-DET and NEU-DET. The quantitative results are presented in Table 4 and Table 5.

As shown in Table 4, our proposed method achieves the highest overall accuracy on the GC10-DET dataset with an mAP of 74.2%. This outperforms both general-purpose baselines (e.g., YOLOv9 at 70.1% and RT-DETR at 69.2%) and specialized defect detectors (e.g., GSCNet at 72.6% and LiFSO-Net at 71.2%). Notably, our method demonstrates substantial gains on challenging categories such as Rp (75.9% AP), showcasing its superior capability in detecting complex multi-scale industrial defects.

Table 4: Comparison with other defect detectors on GC10-DET

| **Model** | **mAP** | **AP**(%) | | | | | | | | | |
|---|---|---|---|---|---|---|---|---|---|---|---|
| | (%) | Pu | Wl | Cg | Ws | Os | Ss | In | Rp | Cr | Wf |
| Yolov8[15] | 67.5 | 99.3 | 90.6 | 96.6 | 76.2 | 57.1 | 65.8 | 30.4 | 27.9 | 55.7 | 74.9 |
| Yolov9[16] | 70.1 | 96.1 | 95.2 | 95.2 | 79.1 | 64.9 | 67.1 | 25.4 | 54.5 | 44.2 | 79.1 |
| RT-DETR[17] | 69.2 | 98.0 | 90.2 | 92.1 | 76.2 | 46.7 | 66.4 | 36.5 | 26.2 | 84.7 | 80.6 |
| Faster R-CNN[3] | 66.7 | 92.8 | 82.4 | 94.5 | 79.9 | 69.1 | 56.0 | 20.6 | 56.6 | 40.3 | 74.6 |
| Dcc-CenterNet[18] | 61.9 | 84.1 | 95.5 | 96.2 | 77.3 | 50.9 | 54.8 | 30.2 | 13.9 | 49.9 | 76.6 |
| MSC-DNet[19] | 69.1 | 97.7 | 95.2 | 92.5 | 75.2 | 67.0 | 61.1 | 37.6 | 48.8 | 31.2 | 84.5 |
| EFD-YOLOv4[20] | 54.7 | 96.3 | 98.0 | 85.3 | 75.0 | 53.3 | 43.2 | 18.2 | 50.0 | 27.3 | 0.0 |
| TCPNet[21] | 70.8 | 99.3 | 92.3 | 94.9 | 79.8 | 49.7 | 68.2 | 38.3 | 28.2 | 85.4 | 83.1 |
| Rbi-YOLOX[22] | 70.5 | 99.0 | 92.5 | 94.2 | 79.6 | 49.5 | 67.8 | 37.6 | 28.3 | 85.3 | 84.8 |
| STFE-Net[23] | 70.6 | 99.5 | 93.1 | 96.9 | 79.5 | 60.9 | 69.4 | 40.7 | 28.0 | 58.2 | 80.0 |
| LiFSO-Net[24] | 71.2 | 99.3 | 94.1 | 91.8 | 79.1 | 46.3 | 62.2 | 47.1 | 34.5 | 70.0 | 87.3 |
| GSCNet[27] | 72.6 | 99.5 | 92.1 | 94.6 | 82.5 | 50.3 | 69.7 | 37.8 | 28.6 | 85.9 | 83.7 |
| Yolov9-e[16] | **71.7** | 95.3 | 92.7 | 94.3 | 79.1 | 75.8 | 69.4 | 39.0 | 48.0 | 51.1 | 72.6 |
| **Our** | **74.2** | 95.4 | 94.6 | 95.7 | 77.7 | 71.7 | 61.2 | 31.4 | 75.9 | 53.9 | 84.6 |

Table 5 summarizes the performance on the NEU-DET dataset, where our method achieves **79.8% mAP**, delivering a **1.1 point gain** over the YOLOv9-e model at 78.7%.While a slight performance gap exists compared to certain customized architectures (e.g., GSCNet), this primarily stems from our deliberate focus: rather than modifying network structures with specialized feature extraction or fusion modules, we strictly retain the unmodified, general-purpose YOLOv9 baseline to center our research on data-efficient training. Although this reveals the architectural limitations of general baselines in specialized scenarios, the fact that our method yields noticeable accuracy improvements purely through theoretical optimization, without adding parameter or computational overhead.

Table 5: Comparison with other defect detectors on NEUDET

| **Model** | **mAP** | **AP(%)** | | | | | |
|---|---|---|---|---|---|---|---|
| | (%) | Cr | In | Pa | Ps | RiS | Sc |
| YOLOv8[15] | 76.8 | 46.4 | 80.4 | 92.4 | 78.9 | 67.3 | 95.1 |
| RT-DETR[17] | 78.6 | 54.3 | 79.6 | 91.2 | 90.6 | 64.2 | 92.1 |
| Faster R-CNN[3] | 77.6 | 32.6 | 82.9 | 91.2 | 92.2 | 69.7 | 96.9 |
| DEA-RetinaNet[26] | 79.1 | 60.9 | 82.5 | 94.3 | 95.8 | 67.2 | 74.1 |
| CABF-FCOS[25] | 76.7 | 55.4 | 75.0 | 93.5 | 88.9 | 62.9 | 84.4 |
| DsP-YOLO[28] | 80.4 | 54.5 | 84.0 | 95.0 | 82.1 | 72.7 | 94.1 |
| DCC-CenterNet[18] | 79.4 | 45.7 | 85.1 | 90.6 | 82.5 | 76.8 | 95.8 |
| MSC-DNet[19] | 79.4 | 42.4 | 84.5 | 94.3 | 91.5 | 71.6 | 92.0 |
| EFD-YOLOv4[20] | 79.9 | 45.8 | 85.4 | 97.0 | 85.0 | 72.8 | 93.6 |
| TCPNet[21] | 82.7 | 93.3 | 57.3 | 92.4 | 85.6 | 94.1 | 69.1 |
| Rbi-YOLOX[22] | 81.0 | 90.2 | 54.3 | 90.0 | 81.3 | 94.2 | 66.4 |
| STFE-Net[23] | 79.2 | 50.4 | 83.3 | 91.5 | 87.0 | 66.9 | 95.9 |
| LiFSO-Net[24] | 79.2 | 48.1 | 86.4 | 94.2 | 88.0 | 64.4 | 94.2 |
| GSCNet[27] | 84.6 | 61.8 | 86.0 | 93.3 | 92.3 | 70.3 | 96.1 |
| Yolov9-e[16] | 78.7 | 40.7 | 84.2 | 92.9 | 87.9 | 69.7 | 96.7 |
| **Our** | 79.8 | 45.9 | 87.6 | 93.4 | 86.5 | 69.9 | 95.6 |

## 6. Conclusions

In this paper, we first theoretically analyzed the intrinsic limitations of conventional IoU-based positive sample selection strategies. Based on this analysis, we designed multiple morphological feature similarity metrics and theoretically examined their respective response functions. Our results demonstrate that incorporating scale and morphological feature similarities can effectively improve the precision of the matching function's response and significantly compress the high-response regions, thereby substantially enhancing the accuracy of the positive sample set. This finding confirms that optimizing the positive sample selection strategy can lead to a significant boost in detection performance for multi-scale and morphologically diverse defect instances, all while incurring zero inference overhead. In future work, we plan to explore richer representations within the feature space to uncover more information beneficial for computing sample-to-instance similarity, aiming to develop more intelligent and information-driven high-level sample selection paradigms for highly efficient industrial visual inspection applications.

## Acknowledgements

This work is supported by National Natural Science Foundation of China (52275527); National Key R&D Program Project (2023YFF0722700,

2024YFF0726603, 2025YFC2428801); Heilongjiang Spring Swan Support Program (CYQN24053).

## Data availability statement

All data included in this study are available upon request by contacting the corresponding author.

## Appendix

## I. Analysis of Conditions for the Existence of Insensitive Regions in IoU Curves

Based on the modeling in Eq. (1), we obtain the start–end coordinate representation of the bounding box:

$$B_{cor} = (x, y, x + \sqrt{area * wh_ratio}, y + \sqrt{area / wh_ratio}) \quad (9)$$

For a candidate box cluster sharing the same starting point, the variation in the IoU curves across different candidate branches changes as a function of the end coordinates.The emergence of the IoU non-sensitive region stems from a decreasing or zero rate of change in the IoU intersection term. A special case occurs when the non-sensitive interval approaches zero, where the end coordinates of a candidate box nearly coincide with those of the ground-truth box, causing the non-sensitive region to appear obscured. As shown in Fig.15 (a-2) and (b-2), both branches contain a candidate box whose end point coincides with that of the ground-truth box. However, the IoU curves of other *wh_ratio* or *area* branches within this candidate box cluster all exhibit insensitive regions, as illustrated in Figures (a-1), (a-3), (b-1), and (b-3). This manifests as none of the end coordinates of the candidate boxes coinciding with the end point of the ground-truth box. Additionally, Table 6 summarizes the parameter representations of the bounding boxes at the onset of the insensitive region for each figure. For a given ground-truth box, within each candidate box cluster, it is only possible to find exactly one candidate box whose end point coincides with that of the ground-truth box in the specific *wh_ratio* and *area* branches, respectively. Therefore, the disappearance of the IoU insensitive region is an extremely low-probability event.

Table 6: Parameter representations of Fig.15

| Index | Ground-Truth Box (red) | Candidate boxes (blue) |
|---|---|---|
| (a-1) | | $B_{can}$=(20,20,100/60,$90^2$/(100/60)) |
| (a-2) | | $B_{can}$=(20,20,90/60,$90^2$/(90/60)) |
| (a-3) | $B_{gt}$=(30,30,1.6,4000) | $B_{can}$=(20,20,80/60,$80^2$/(80/60)) |
| (b-1) | | $B_{can}$=(20,20, $60^2$/5000,5000) |
| (b-2) | | $B_{can}$=(20,20,90/60,5400) |
| (b-3) | | $B_{can}$=(20,20, $90^2$/5800,5800) |

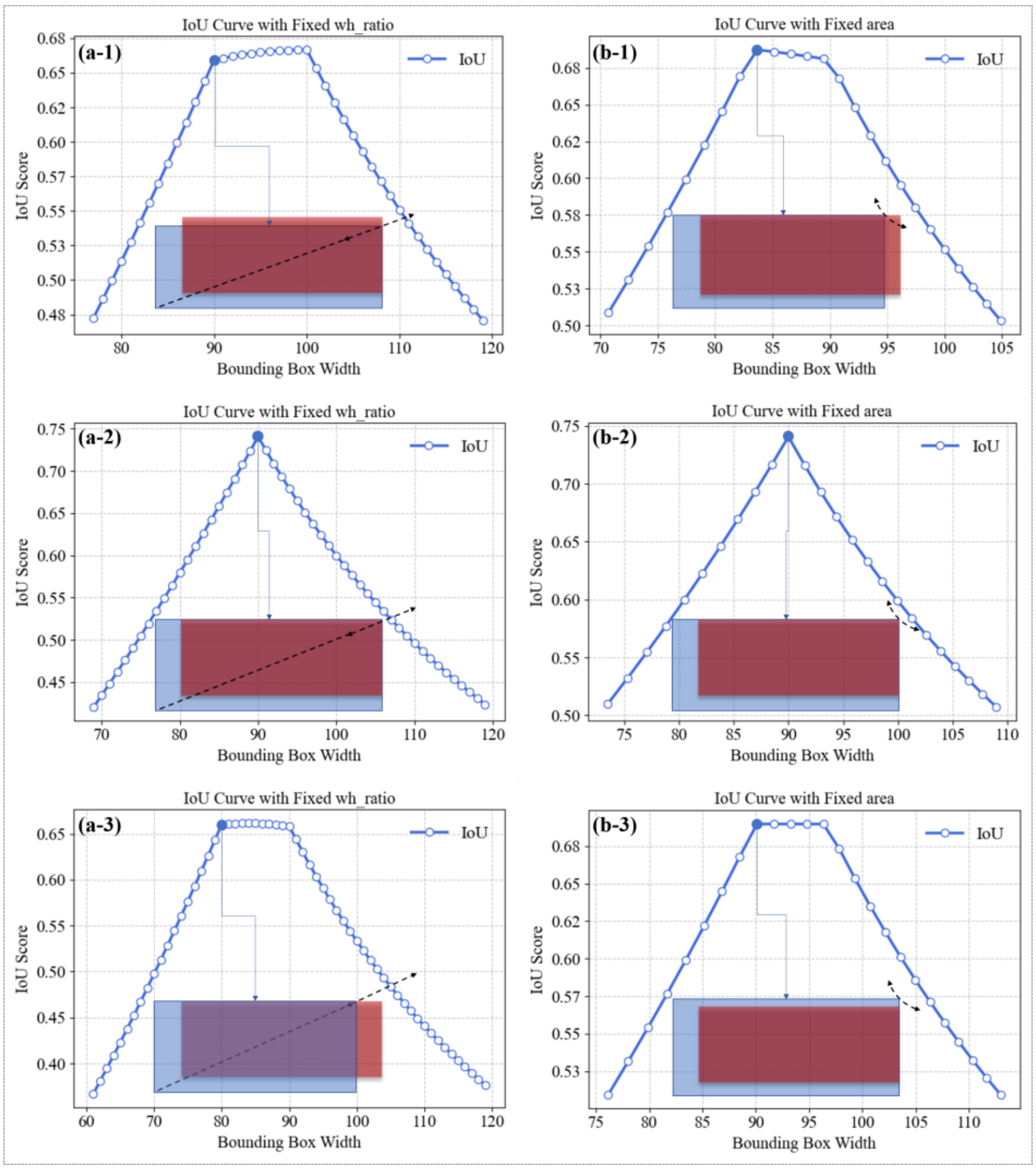

Fig.15 Visualization analysis of IoU curves across different branches for the candidate box cluster starting at (20, 20). (Arrows indicate variation trends)